\documentclass[10pt,twocolumn,letterpaper]{article}
\usepackage{iccv}
\usepackage{times}
\usepackage{epsfig}
\usepackage{graphicx}
\usepackage{amsmath}
\usepackage{amssymb}
\usepackage{multirow}
\usepackage{gensymb}

\usepackage{balance}
\usepackage{ctable}
\usepackage{stfloats}
\usepackage{array}
\usepackage{float}
\newcolumntype{Q}{>{\centering\arraybackslash}m{1cm}}
\newcolumntype{O}{>{\centering\arraybackslash}m{0.8cm}}
\newcolumntype{V}{>{\centering\arraybackslash}m{1.8cm}}
\newcolumntype{R}{>{\centering\arraybackslash}m{0.9cm}}
\newcolumntype{L}{>{\centering\arraybackslash}m{1.5cm}}
\newcolumntype{S}{>{\raggedright\arraybackslash}m{3.8cm}}

\usepackage[pagebackref=true,breaklinks=true,letterpaper=true,colorlinks,bookmarks=false]{hyperref}

\iccvfinalcopy 

\def\iccvPaperID{5036} 

\ificcvfinal\fi
\begin{document}

\title{Direct Image to Point Cloud Descriptors Matching\\for 6-DOF Camera Localization in Dense 3D Point Cloud}

\author{Uzair Nadeem$^1$, Mohammad A. A. K. Jalwana$^1$, Mohammed Bennamoun$^1$,\\ Roberto Togneri$^2$, Ferdous Sohel$^3$\\
1 Department of Computer Science and Software Engineering, The University of Western Australia\\
2 Department of Electrical, Electronics and Computer Engineering, The University of Western Australia\\
3 School of Engineering and Information Technology, Murdoch University\\
{\tt\small \{uzair.nadeem@research.,mohammad.jalwana.@research,mohammed.bennamoun@,roberto.togneri@\}}\\{\tt\small uwa.edu.au,f.sohel@murdoch.edu.au}
}

\maketitle

\begin{abstract}

\vspace{-0.7cm}We propose a novel concept to directly match feature descriptors extracted from RGB images, with feature descriptors extracted from 3D point clouds. We use this concept to localize the position and orientation (pose) of the camera of a query image in dense point clouds. We generate a dataset of matching 2D and 3D descriptors, and use it to train a proposed Descriptor-Matcher algorithm. To localize a query image in a point cloud, we extract 2D keypoints and descriptors from the query image. Then the Descriptor-Matcher is used to find the corresponding pairs 2D and 3D keypoints by matching the 2D descriptors with the pre-extracted 3D descriptors of the point cloud. This information is used in a robust pose estimation algorithm to localize the query image in the 3D point cloud. Experiments demonstrate that directly matching 2D and 3D descriptors is not only a viable idea but also achieves competitive accuracy compared to other state-of-the-art approaches for camera pose localization.    
\end{abstract}

\begin{figure*}[b]
\begin{center}
   \includegraphics[width=1\linewidth]{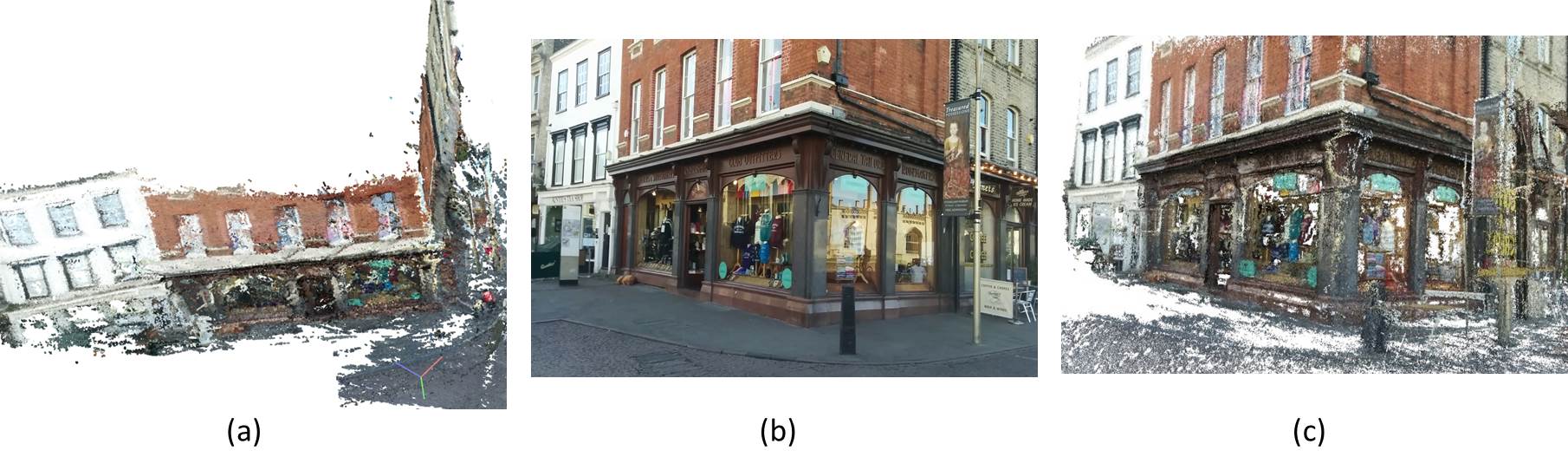}
\end{center}
   \caption{(a) A section of the 3D point cloud from Shop Facade dataset \cite{posenet}. (b) An RGB query image to be localized in 3D point cloud (c) Visualization of the area of the 3D point cloud, identified by our technique as the location of the query image.}
\label{fig:result}
\end{figure*}

\section{Introduction}
With the recent advances in the computational power of modern machines, 3D vision and its applications are becoming more and more popular. A lot of research is being carried out using RGB-D images, point clouds and 3D meshes.This is also supported by new hardware and software products that use 3D data for various applications. For example, Microsoft Windows 10 has now a dedicated library for point clouds and meshes named ``3D Objects", as well as 3D Paint and 3D Builder software. The quality and capabilities of 3D sensors have also improved a lot, and at the same time, costs have come down. Many of the latest smart-phones have a dedicated 3D sensor along with one or more RGB cameras. 

Despite the progress in the fields of 3D and 2D vision, there is still a need to develop better techniques for the fusion of 2D and 3D information. Both 2D and 3D data can complement each others in many ways for improved performance and efficiency. Techniques that can combine 2D and 3D data have many potential applications, e.g., recognizing an object in an image using its 3D model, locating objects or regions of interest from RGB images in 3D maps, face recognition and verification, as well as 2D image localization in a 3D point cloud. Moreover, most of the vision data that is currently available is in the form of 2D images (or videos). A technique that can bridge the gap between the 2D and 3D data will highly be beneficial to improve 3D vision systems using 2D data. In this paper we propose a technique to match feature descriptors extracted from 2D images and 3D point clouds and demonstrate their application for the task of 6-DOF camera pose (position and orientation) estimation of 2D images in a 3D point cloud.

Camera pose estimation is a popular research topic now-a-days because of its applications e.g., for place recognition, augmented reality, robotic navigation, robot pose estimation, grasping, and in Simultaneous Localization and Mapping (SLAM) problem. Current techniques for camera localization comprise two major categories: \textbf{(ii)} Regression Networks-based methods and \textbf{(i)} Handcrafted features-based methods. The regression networks-based methods usually require a lot of data for training and have high requirements for computational resources, such as powerful GPUs. Most of the feature-based techniques use local or global features that are extracted from 2D images. For 3D models they use sparse 3D point clouds generated from the Structure from Motion (SfM) pipeline \cite{colmapsfm}. Once the 3D model is generated from SfM, the approach provides a one-to-one correspondence between the pixels of the images and the 3D points in the SfM model. This correspondence has been used to create systems that can localize images in the sparse point cloud, generated by SfM. However, with the advances in 3D scanning technology, it is possible to directly generate 3D point clouds of large areas without using the SfM pipeline e.g., Microsoft Kinect, LIDAR or Faro 3D scanner or Matterport scanners can be used to directly generate the point cloud of a scene. This is a much more practical approach to generate 3D models, compared to SfM as the generated point clouds are denser and the quality of the models generated by 3D scanners is much better than the ones obtained by SfM. Although it may be possible to generate sparse point clouds from the dense ones by randomly downsampling them, it is not possible to generate one-to-one correspondences between the 3D vertices and the pixels of the 2D images. In such a scenario, many of the current image localization techniques cannot be used since they heavily rely on the information that is generated by the SfM.

\begin{figure*}[t]
\begin{center}
   \includegraphics[width=1\linewidth]{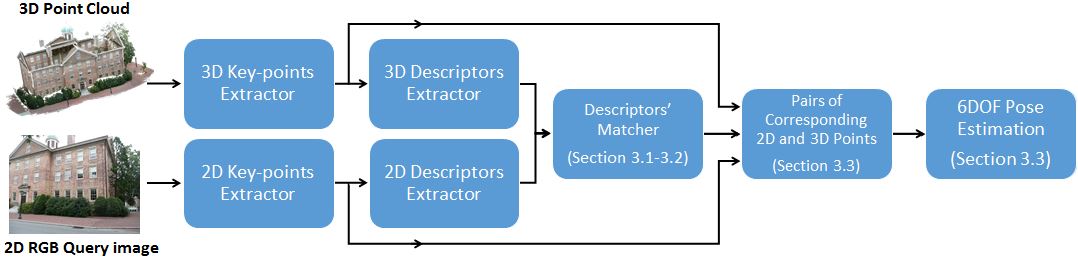}
\end{center}
   \caption{A block diagram of the test pipeline of the proposed technique. We extract 3D keypoints and descriptors from the dense 3D Point Cloud. 2D keypoints and descriptors are extracted from the 2D RGB Query Image. Then our proposed ``Descriptor Matches" algorithm directly matches the 2D descriptors with the 3D descriptors to generate correspondence between points in 2D image and 3D point cloud. This is then used with a robust pose estimation algorithm to estimate the 6-DOF pose of the query image in the 3D point cloud.}
\label{fig:test_chart}
\end{figure*}

In this paper we propose a novel camera localization technique. Our technique can localize images directly in dense point clouds, that are acquired by 3D scanners, by directly matching the 2D descriptors extracted from images with the 3D descriptors that are extracted from the point clouds. Figure \ref{fig:result} shows the dense point cloud of the Shop Facade dataset \cite{posenet} along with a query image localized by our technique in the point cloud. Specifically, we train a 2D-3D Descriptor matching classifier, that we call `Descriptor-Matcher', using a training set of corresponding 2D and 3D descriptors. To localize an image, we extract keypoints and descriptors from the query image using 2D keypoints and descriptors techniques such as Scale Invariant Features Transform (SIFT) keypoints and SIFT descriptors \cite{lowesift}. Similarly, we extract 3D keypoints and descriptors from the point cloud e.g., 3D-SIFT keypoints \cite{lowesift, pcl} or 3D-Harris keypoints \cite{pcl, harris} and Rotation Invariant Features Transform (RIFT) descriptors \cite{rift}. Then we use the trained Descriptor-Matcher to find the matching between 2D and 3D descriptors. We use this information to identify the points from the image that match with points in the 3D point cloud. The matching pairs of 2D and 3D points are used in a robust pose estimation algorithm to estimate the orientation and the location of camera in the 3D point cloud (see Section \ref{technique:test}). Figure \ref{fig:test_chart} shows the steps of the proposed technique to estimate the camera pose.

To the best of our knowledge, our contribution is the first: \textbf{(i)} to directly match 2D descriptors from RGB images with 3D descriptors that are extracted from dense point clouds, and \textbf{(ii)} to localize 6-DOF Camera pose in dense 3D point clouds by directly matching 2D and 3D descriptors. 

The rest of this paper is organized as follows. Section \ref{related_work} discusses the different types of techniques that have the goal to loacalize or geo-register an image. The proposed technique is presented in Section \ref{technique}. Section \ref{experiments} reports the experimental setup and a detailed evaluation of our results along with a comparison with other approaches. The paper is concluded in Section \ref{conclusion}.

\section{Related Work}\label{related_work}








Localizing the position of an image and estimating its pose is an active area of research. Two major approaches are used for this purpose: \textbf{(i)} handcrafted features-based methods, and \textbf{(ii)} network-based pose regression.

\textbf{(i) Handcrafted features-based methods} mainly emphasize on locating an image in an area i.e., many of the image localization datasets do not provide the ground truth camera pose, therefore, the localization techniques only use the number of inliers found with RANSAC \cite{ransac} as a registration metric. For example, if the RANSAC finds 12 inliers for the query image, the image is considered registered. This is a very crude metric and does not actually show the performance of these techniques. These works are further of two types, \textbf{(i.a)} image retrieval based methods, and \textbf{(i.b)} SfM-based methods.

\textbf{(i.a) Image retrieval based methods } treat the problem as a pure image based problem where for a given query image, retrieval techniques are used to find a set of similar images from a database of geo-tagged images. This set of images is then used to triangulate the pose of the camera \cite{chen2011city, zamir2010accurate,zhang2006image}. As evident by the approach, they cannot be used to localize images in a 3D model. Also since the estimated pose is based on the poses of the retrieved similar images, which are not necessarily close, so this approximation is inaccurate compared to the techniques that use the SfM model for localization \cite{sattler2017efficient}.

\textbf{ (i.b) SfM-based methods} use structure from motion based 3D models for localization. These methods are more accurate than the image retrieval based methods. In SfM each point is triangulated from multiple images' keypoints and is then associated with those corresponding feature descriptors \cite{colmapsfm}. The commonly used image features include SIFT, SURF, ORB etc. Irschara et al., \cite{irschara2009structure} created the SfM model from the training images and then used image retrieval techniques to find training images similar to the query image. The retrieved images were used with the SfM model to improve the localization accuracy. Yunpeng et al., \cite{li2010location} further enhanced the performance by directly comparing the query image and 3D point features, instead of using retrieval methods. Sattler et al., \cite{sattler2015hyperpoints} used a visual vocabulary of 16M words for localization in larger point clouds. Sattler et al., \cite{sattler2017efficient} extended their work by using a prioritized matching system to improve the image localization results.

These works require that the point cloud be generated from the SfM based techniques, due to the fact that 3D points are assigned features based on the known poses of images that constructed them. As each point in the point cloud generated from SfM is created from keypoints that are extracted from a number of 2D images, a function of the 2D descriptors corresponding to those keypoints is considered by some works as 3D descriptors (e.g., average of 2D descriptors). However, such approximation and assignment of 3D descriptors is only possible if the 3D model is generated from the SfM using a large number of images. Moreover, since 3D descriptors are a function of the 2D image descriptors, the matching between image descriptors and SfM model's descriptors is not a 2D to 3D matching problem in the true sense, but rather a 2D to 2D descriptor matching problem. These algorithm suffer from the inherent issues of SfM i.e., the images need to have sufficient texture and good quality in order to get localized. The recent arrival of cheap 3D sensors allows directly capture large scale point clouds avoiding the time consuming generation of 3D models from hundred or thousands of images. The above mentioned techniques cannot localize images in the point clouds captured by 3D scanners.

\textbf{(ii) Network-based pose regression methods} use neural networks for pose estimation. PoseNet, developed by Kendall et al. \cite{posenet}, is a CNN based neural network which learns to regress input image to the pose. However, it has a low accuracy. Kendall et al., \cite{kendall2017geometric} and Walch et al., \cite{walch2017image} enhanced Posenet by improving the loss function and architecture of the network, respectively. Instead of directly regressing the pose, Brachmann et al. \cite{brachmann2017dsac} introduced the differentiable RANSAC for using it as a trainable component in their deep network pipeline based on the work of \cite{shotton2013scene}. However, the localization results of the network-based methods are generally inferior to the SfM based techniques \cite{brachmann2018learning}.

In contrast to SfM-based techniques, our technique is the first approach which extracts 3D keypoints and descriptors from the point cloud and matches them with the features extracted from the 2D images. This allows our technique to find the pose of the query image in the 3D model that can be obtained from any scanner.



\section{Technique}\label{technique}
The proposed technique provides a novel way to estimate poses of query images in a 3D point cloud. Our technique can work with point clouds that are acquired from any scanner. Moreover, our technique can work in both indoor and outdoor scenarios. As no popular dataset for pose estimation or image localization provides dense 3D point clouds and poses to train and query images using the the dense model, we create datasets by using Multi View Stereo (MVS). Specifically, we used the COLMAP's MVS \cite{colmapmvs} pipeline to generate dense 3D point clouds from images in publicly available datasets.  

To train the Descriptor-Matcher, we need a dataset of pairs of 2D and 3D descriptors which correspond to the same point in the 2D image and the 3D point cloud, respectively. Once the Descriptor-Matcher is trained, our technique does not require any type of intermediate information (e.g., from SfM) to localize the query image and estimate its pose.  

\subsection{Dataset Creation}
\label{technique:dataset}
To train the Descriptor-Matcher to match the 2D descriptors with the 3D descriptors, we need a dataset of corresponding 2D and 3D descriptors. As any manual creation of such a dataset on a large scale will be tedious and impractical, we devised a way to automatically mine matching 2D and 3D descriptors from a given set of training images, and a dense 3D point cloud that may have been generated from any scanner. Given a dataset with $N$ number of training images and $M$ number of query images, we use the COLMAP's MVS \cite{colmapmvs} to generate the dense 3D point clouds. COLMAP generates a sparse SfM model \cite{colmapsfm} (called sparse point cloud) in an intermediate step to the construction of the dense point cloud. This provides the poses of the training and the query images, with respect to the sparse 3D point cloud. For the quantitative evaluation of the proposed technique, we need the poses of the query images with respect to the point cloud obtained from 3D scanner (called the Dense 3D point cloud). The sparse point cloud can be registered with point clouds obtained from any 3D scanner, to get the poses of training and query images with respect dense 3D point cloud. For our experiments we used the point cloud generated from MVS as the dense point cloud. 

For the dataset creation, we first extract the 3D keypoints and descriptors from the dense point cloud \cite{pcl}. Similar to \cite{li2010location, sattler2017efficient, sattler2018benchmarking}, we remove all the points in the point cloud that were generated using only the query images to ensure that no information from the query set is used for the generation of the training dataset. After applying the 3D keypoint detector and descriptors on the dense point cloud we get a set of 3D keypoints:

\begin{equation}
\label{e1}
keys_{3D}= \bigcup_{i=1}^{N_{k3}}(x_i,y_i,z_i)
\end{equation}

and their corresponding 3D descriptors:

\begin{equation}
\label{e2}
desc_{3D}= \bigcup_{i=1}^{N_{k3}}(d_i^1,d_i^2,d_i^3,...d_i^{m})
\end{equation}

where $N_{k3}$ is the number of detected keypoints and $m$ is the length of the used 3D descriptor. Let $PC_s$ be the sparse point cloud obtained after removing the points that were created using the query images:

\begin{equation}
\label{e3}
PC_{s}= \bigcup_{j=1}^{N_{p}}(x_j,y_j,z_j)
\end{equation}

where $N_p$ is the number of points that are left in the point cloud. We find the Euclidean distances between the elements of $keys_{3D}$ and $PC_{s}$ and keep only those points whose difference is less than a threshold $\alpha$. Let $\zeta$ be the set of the pair of indices, corresponding to the points of $N_{k3}$ and $PC_{s}$, that are equal within an error threshold $\alpha$ :

 \begin{equation}\label{e4}
  \begin{aligned}
\zeta= & \bigcup(i,j) \quad \forall  \left || (x_i-x_j),(y_i-y_j),(z_i-z_j)) \right || < \alpha \\ \quad  & where \quad i=1,2,3,...,N_{k3}, \quad  j=1,2,3,...,N_p. 
 \end{aligned}
\end{equation}

It is to be noted that the resulting $\zeta$ usually contains less than $10\%$ of the keypoints that are detected from the dense point cloud. The fact that our technique produces good results on the query images (see Section \ref{experiments}), despite never being trained on more than 90\% of the training 3D keypoints, shows the high generalization capability of the proposed technique. Since the sparse point cloud, created by COLMAP \cite{colmapsfm}, contains the correspondence information between the pixels from the training images that were used to create 3D points in the sparse point cloud, we use this information along with  $\zeta$ to find the 2D keypoints corresponding to the keypoints obtained from the dense point cloud which are included in the set $\zeta$. This creates a correspondence between the 3D keypoints from the dense point cloud and the 2D keypoints from the training images. Note that this is a one-to-many correspondence, since one 3D point in a point cloud may appear in many 2D images. We convert this set of keypoints correspondences from one-to-many to one-to-one by duplicating each 3D keypoint as many times as their are corresponding 2D keypoints for that point. This gives us a one-to-one set of matching 2D and 3D keypoints. We retrieve the corresponding descriptors for 3D keypoints from $desc_{3D}$ and apply 2D descriptor extraction on the 2D keypoints to obtain a dataset of corresponding 2D and 3D descriptors.

\begin{table*}[h]
\label{tab:southb}
  \centering
    \begin{tabular}{|l|c|c|c|c|c|c|}
    \hline
    \textbf{Errors \textbackslash{} Metrics} & \textbf{Median} & \textbf{P 25\%} & \textbf{P 50\%} & \textbf{P 75\%} & \textbf{P 90\%} & \textbf{P 95\%} \\
    \hline
    \textbf{Position Error (m)} & 0.12 m & 0.086 m & 0.12 m & 0.15 m & 0.21 m & 0.48 m \\
    \hline
    \textbf{Angle Error (degrees)} & 0.61\degree & 0.36\degree & 0.61\degree & 0.93\degree & 1.26 \degree & 1.42\degree \\
    \hline
    \end{tabular}%
    \caption{Median and percentile localization errors for the South Building dataset \cite{colmapsfm,colmapmvs}. P stands for percentile, e.g., P 25\% means the maximum error for 25\% of the data when errors are sorted in the ascending order.}
\end{table*}%

\begin{table*}[h]
\label{tab:notredame}
  \centering
    \begin{tabular}{|l|c|c|c|c|c|c|}
    \hline
    \textbf{Errors \textbackslash{} Metrics} & \textbf{Median} & \textbf{P 25\%} & \textbf{P 50\%} & \textbf{P 75\%} & \textbf{P 90\%} & \textbf{P 95\%} \\
    \hline
    \textbf{Position Error (m)} & 4.04 m & 2.15 m & 4.04 m & 15.66 m & 33.59 m & 62.56 m \\
    \hline
    \textbf{Angle Error (degrees)} & 4.10\degree  & 2.18\degree & 4.10\degree & 14.63\degree & 27.60\degree & 66.90\degree \\
    \hline
    \end{tabular}%
    \caption{Median and percentile localization errors for the Notre Dame dataset \cite{notredame} . P stands for percentile, e.g., P 25\% means the maximum error for 25\% of the data when errors are sorted in the ascending order.}
\end{table*}%

\begin{table*}[h]
\label{tab:shop}
  \centering
    \begin{tabular}{|l|c|c|c|c|c|c|}
    \hline
    \textbf{Errors \textbackslash{} Metrics} & \textbf{Median} & \textbf{P 25\%} & \textbf{P 50\%} & \textbf{P 75\%} & \textbf{P 90\%} & \textbf{P 95\%} \\
    \hline
    \textbf{Position Error (m)} & 0.23 m & 0.09 m & 0.23 m & 1.73 m & 11.07 m & 17.31 m \\
    \hline
    \textbf{Angle Error (degrees)} & 1.60\degree & 0.67\degree & 1.60\degree & 15.08\degree & 86.72\degree & 127.28\degree \\
    \hline
    \end{tabular}%
    \caption{Median and percentile localization errors for the Shop Facade dataset from the Cambridge Land Marks database \cite{posenet}. P stands for percentile, e.g., P 25\% means the maximum error for 25\% of the data when errors are sorted in the ascending order.}
\end{table*}%

\begin{table*}[hb]
\label{tab:hospital}
  \centering
    \begin{tabular}{|l|c|c|c|c|c|c|}
    \hline
    \textbf{Errors \textbackslash{} Metrics} & \textbf{Median} & \textbf{P 25\%} & \textbf{P 50\%} & \textbf{P 75\%} & \textbf{P 90\%} & \textbf{P 95\%} \\
    \hline
    \textbf{Position Error (m)} & 0.65 m & 0.28 m & 0.65 m & 12.16 m & 34.42 m & 42.78 m \\
    \hline
    \textbf{Angle Error (degrees)} & 1.14\degree & 0.61\degree & 1.14\degree & 47.17\degree & 105.30\degree & 129.24\degree \\
    \hline
    \end{tabular}%
    \caption{Median and percentile localization errors for the Old Hospital dataset from the Cambridge Land Marks database \cite{posenet}. P stands for percentile, e.g., P 25\% means the maximum error for 25\% of the data when errors are sorted in the ascending order.}
\end{table*}%

\begin{table*}[hb]
\label{tab:heads}
  \centering
    \begin{tabular}{|l|c|c|c|c|c|c|}
    \hline
    \textbf{Errors \textbackslash{} Metrics} & \textbf{Median} & \textbf{P 25\%} & \textbf{P 50\%} & \textbf{P 75\%} & \textbf{P 90\%} & \textbf{P 95\%} \\
    \hline
    \textbf{Position Error (m)} & 0.01 m & 0 .008 m & 0.01 m & 0.09 m & 0.67 m & 0.93 m \\
    \hline
    \textbf{Angle Error (degrees)} & 1.82\degree & 0.84\degree & 1.82\degree & 7.43\degree & 100.48\degree & 129.75\degree \\
    \hline
    \end{tabular}%
    \caption{Median and percentile localization errors for the Heads dataset from the the Seven Scenes RGBD database \cite{sevenscenes}. P stands for percentile, e.g., P 25\% means the maximum error for 25\% of the data when errors are sorted in the ascending order.}
\end{table*}%
\begin{table*}[htbp]
\label{tab:fire}
  \centering
    \begin{tabular}{|l|c|c|c|c|c|c|}
    \hline
    \textbf{Errors \textbackslash{} Metrics} & \textbf{Median} & \textbf{P 25\%} & \textbf{P 50\%} & \textbf{P 75\%} & \textbf{P 90\%} & \textbf{P 95\%} \\
    \hline
    \textbf{Position Error (m)} & 0.02 m & 0.006 m & 0.02 m & 0.45 m & 1.06 m & 1.52 m \\
    \hline
    \textbf{Angle Error (degrees)} & 1.56\degree & 0.86\degree & 1.56\degree & 50.44\degree & 147.05\degree & 162.42\degree \\
    \hline
    \end{tabular}%
    \caption{Median and percentile localization errors for the Fire dataset from the the Seven Scenes RGBD database \cite{sevenscenes}. P stands for percentile, e.g., P 25\% means the maximum error for 25\% of the data when errors are sorted in the ascending order.}
\end{table*}%
\begin{table*}[htbp]
\label{tab:pumpkin}
  \centering
    \begin{tabular}{|l|c|c|c|c|c|c|}
    \hline
    \textbf{Errors \textbackslash{} Metrics} & \textbf{Median} & \textbf{P 25\%} & \textbf{P 50\%} & \textbf{P 75\%} & \textbf{P 90\%} & \textbf{P 95\%} \\
    \hline
    \textbf{Position Error (m)} & 0.01 m & 0.007m  & 0.01 m & 0.02 m & 1.93 m & 3.76 m \\
    \hline
    \textbf{Angle Error (degrees)} & 0.66\degree & 0.40\degree & 0.66\degree & 1.30\degree & 131.23\degree & 178.37\degree \\
    \hline
    \end{tabular}%
    \caption{Median and percentile localization errors for the Pumpkin dataset from the the Seven Scenes RGBD database \cite{sevenscenes}. P stands for percentile, e.g., P 25\% means the maximum error for 25\% of the data when errors are sorted in the ascending order.}
\end{table*}%
\begin{table*}[htbp]
  \centering
 
    \begin{tabular}{|L|R|R|R|R|R|R|R|R|R|R|R|R|}
    \hline
   \textbf{Methods$\rightarrow$} & \multicolumn{2}{V|}{\textbf{PoseNet \cite{posenet} ICCV'15}} & \multicolumn{2}{V|}{\textbf{Geom. Loss Net \cite{kendall2017geometric} CVPR'17}} & \multicolumn{2}{V|}{\textbf{VLocNet \cite{valada2018deep} ICRA'18}} & \multicolumn{2}{V|}{\textbf{DSAC \cite{brachmann2017dsac} CVPR'17}} & \multicolumn{2}{V|}{\textbf{Active Search \cite{sattler2017efficient} TPAMI'17}} & \multicolumn{2}{V|}{\textbf{Ours}} \\
    \hline
    \textbf{Methods Type$\rightarrow$} & \multicolumn{2}{V|}{\textbf{Network-based}} & \multicolumn{2}{V|}{\textbf{Network-based}} & \multicolumn{2}{V|}{\textbf{Network-based}} & \multicolumn{2}{V|}{\textbf{Network + RANSAC}} & \multicolumn{2}{V|}{\textbf{SfM-based}} & \multicolumn{2}{V|}{\textbf{2D to 3D Descriptors Matching}} \\
    \hline
    \textbf{Datsets$\downarrow$} & \textbf{Pos} & \textbf{Ang} & \textbf{Pos} & \textbf{Ang} & \textbf{Pos} & \textbf{Ang} & \textbf{Pos} & \textbf{Ang} & \textbf{Pos} & \textbf{Ang} & \textbf{Pos} & \textbf{Ang} \\
    \hline
    \textbf{Shop Facade} & 1.46 m & 4.04\degree  & 1.05 m & 4\degree   & 0.59 m & 3.529\degree  & 0.09 m & 0.4\degree    & 0.12 m & 0.4\degree    & 0.23 m & 1.6\degree  \\
    \hline
    \textbf{Old Hospital} & 2.31 m & 2.69\degree   & 2.17 m & 2.9\degree    & 1.07 m & 2.411\degree  & 0.33 m & 0.6\degree    & 0.44 m & 1\degree      & 0.65 m & 1.14\degree  \\
    \hline
    \textbf{Heads} & 0.29 m & 6\degree     & 0.17 m & 13\degree     & 0.05 m & 6.64\degree   & 0.03 m & 2.7\degree    & 0.02 m  & 1.5\degree    & 0.01 m & 1.82\degree  \\
    \hline
    \textbf{Fire } & 0.47 m & 7.33\degree   & 0.27 m & 11.3\degree   & 0.04 m & 5.34\degree   & 0.04 m & 1.5\degree    & 0.03 m & 1.5\degree    & 0.02 m & 1.56\degree  \\
    \hline
    \textbf{Pumpkin} & 0.47 m & 4.21\degree   & 0.26 m & 4.8\degree    & 0.04 m & 2.28\degree   & 0.05 m & 2\degree      & 0.08 m & 3.1\degree    & 0.01 m & 0.66\degree  \\
    \hline
    \end{tabular}%
     \caption{Median errors for position and orientation estimation of our technique compared to other approaches on indoor and outdoor datasets. Pos stands for median positional error in meters and Ang stands for rotational error in degrees.}
  \label{tab:comparison}%
\end{table*}%

\subsection{Training}
\label{technique:train}
We use the generated dataset to train the Descriptor-Matcher. We pose the problem of matching 2D and 3D descriptors as a binary classification problem. Empirically, we found that the Random Forest classifier \cite{randomforest} is the most suited for this problem, due to its robustness, speed, resistance to over-fitting and good generalization ability. We concatenated the corresponding 3D and 2D descriptors from the training dataset and used them as positive samples. To generate negative samples, we generate one-to-one matches between the 3D and the 2D descriptors from the training set, such that the distance between their corresponding keypoints is greater than a threshold $\beta$. Due to the large number of negative samples, we randomly select a small subset of them. We used the positive and the negative samples to train our Descriptor-Matcher.

\subsection{Testing}
\label{technique:test}
To localize a given query image at test time, we extract 2D keypoints and descriptors from the query image. Similarly, we extract 3D keypoints and descriptors from the point cloud in which the query image is to be localized. We then use the Descriptor-Matcher to find the matching 2D and 3D descriptors. We retrieve the corresponding 2D and 3D keypoints of the matching 2D and 3D descriptors based on the output of the Descriptor-Matcher and use them as the matching pairs of points in the 2D image and the 3D point cloud. Just like any other classifier, there is the possibility of false positives in the resulting matches. We therefore use a robust pose estimation algorithm to estimate the location and orientation of the 2D image. Our pose estimation algorithm is a combination of P3P \cite{p3pmatlab} and MLESAC\cite{mlesacmatlab} algorithms. MLESAC is a generalized and improved version of RANSAC\cite{ransac} that not only tries to maximize the number of inliers but also maximizes the likelihood\cite{mlesacmatlab}. This provides us with the position and viewing direction of the camera used to capture the query image in the 3D point cloud. We can use this position and orientation to map the image on the 3D point cloud. Figure \ref{fig:test_chart} shows the testing pipeline of the proposed technique.

\section{Experiments and Analysis}\label{experiments}
To evaluate the performance of our technique, we carried out experiments on publicly available popular datasets. As none of these datasets contains dense point clouds, we generated them using 2D images from the datasets (see Section \ref{technique:dataset}). We used the Heads, Fire and Pumpkin datasets from the RGB-D 7-Scenes Database \cite{sevenscenes} to evaluate our technique for indoor scenarios. For outdoor scenarios, we used the South Building \cite{colmapsfm,colmapmvs}, Notre Dame \cite{notredame}, Shop Facade \cite{posenet} and Old Hospital \cite{posenet} datasets.

The quality of the point clouds generated from MVS turned out to be poor as they contained a large amount of noise. To reduce the noise, we used the Statistical Outlier Removal Filter (SOR) \cite{pcl}. SOR computes the mean distance between points in a local neighbourhood, and removes those points that are far apart based on a statistical threshold.

We extracted SIFT keypoints and descriptors \cite{lowesift} from the 2D images and 3D SIFT keypoints \cite{pcl} and RIFT descriptors \cite{rift} from the dense 3D point cloud. 3D SIFT keypoint extraction is an extension of the 2D SIFT keypoint extraction algorithm \cite{lowesift} for point clouds. In order to optimize the Random forest in the Descriptor-Matcher, we held out 15\% of the training dataset (see Secion \ref{technique:dataset}) for the validation set and trained the Random Forest on the remaining 85\% of the data. The validation set was used to optimize the parameters of the the Random Forest. These optimal parameters were then used to train the Random Forest using the complete training dataset. 

\subsection{Evaluation Metric}
We calculated the errors in position and rotation of the query images with respect to the ground truth. For errors in position we used the distance between the ground truth and the predicted positions of the camera in the point cloud for the evaluation metric. For rotational error, we calculated the smallest angle between the viewing direction of the ground truth orientation and the predicted orientation of the query image's camera. If $R_{g}$ is the ground truth rotation matrix and $R_p$ is the predicted rotation matrix then the rotational error $\phi$ in degrees is calculated as follows:
\begin{equation}
\label{eq:R_err}
    \phi = \frac{\pi}{180}*cos^{-1}(\frac{trace(R_g \times transpose(R_p))-1}{2})
\end{equation}

\subsection{Outdoor Datasets}
\subsubsection{South Building}
The South Building dataset \cite{colmapsfm,colmapmvs} consists of images of the “South” building at the university of North Carolina, Chapel Hill. The dataset contains images from all the sides of the building. We randomly selected 10\% of the images for the query set and the remaining images were used for the training set. The images in this dataset were acquired from different viewpoints, orientations, and occlusions were present due to the trees. The South building has also repetitive patterns. Moreover, due to the symmetric nature of the building, it is difficult to differentiate images from different sides of the building. Table 1 shows the results of the proposed technique on the test set. The proposed technique was able to register all the images of the query set. We report the median as well as the percentile errors for the intervals of 25\%, 50\%, 75\%, 90\% and 95\% data, for both position (in meters) and angle (in degrees).

\subsubsection{Notre Dame}
The Notre Dame dataset \cite{notredame} contains random pictures of the Notre Dame Cathedral and its surrounding areas in Paris, France. There is very high variation in the images in terms of scale, illumination and orientation. The dataset contains 716 images. The scale of the pictures varies from a close up of the graffiti on the cathedral to ones taken from hundreds of meters away from the building. The pictures are both in landscape and portrait orientation. We did not correct the orientation of the pictures and used them as they were, for training and testing purposes. Moreover, this dataset contains picture that were taken at different times of the day including day, night and dusk. The pictures taken at night look strikingly different from the ones taken at day time. We randomly selected 10\% of the images as the held out query set and used the remaining pictures for training. Our technique successfully registered 57 query images out of 72. Among the pictures that were not registered, 4 were not part of the 3D Model so these were successful rejections by our technique. Table 2 shows the localization results on the Notre Dame dataset for the registered images.

\subsubsection{Shop Facade}
 The Shop Facade dataset, from the Cambridge Landmarks database \cite{posenet}, consists of 334 images from the intersection of two streets in Cambridge, with a focus on the shops at the intersection. We used the standard train-query split defined by the authors of the dataset. The query set contains 103 images \cite{posenet}. Our technique successfully localized all the images in the query set. Table 3 shows the median errors in position and rotation along with the percentile errors on the Shop Facade dataset.

\subsubsection{Old Hospital}
The Old Hospital dataset is part of the Cambridge Landmarks database \cite{posenet}. This dataset contains 1077 images. The dataset suffers from the challenges of high symmetry and repetitive patterns in the buildings. We used the data split defined by the authors of the dataset \cite{posenet} to generate the training and the query sets. The query set contains 182 images. We were able to localize all the images of the query set in the 3D model. Table 4 shows the percentile errors in the location and the orientation of the localized images for the intervals of 25\%, 50\%, 75\%, 90\% and 95\% data.

\subsection{Indoor Datasets}
To test the performance of the proposed technique in indoor conditions we used the Heads, Pumpkin and Fire datasets. For indoor datasets, the quality of the generated point clouds from MVS was much worse than what would have been achieved, had a Kinect or some other 3D scanner been used. We only applied SOR filter to the generated point cloud and used it for our experiments. The results of our technique on the indoor datasets demonstrate its robustness against noise in point clouds.

\subsubsection{Heads}
The Heads dataset \cite{sevenscenes} for indoor localization is part of the seven scenes RGBD database. As the name indicates, it contains multiple models of heads, along with monitors, and other items in a room. The dataset consists of 2000 images. We used the standard train-query split as defined by the authors of the dataset \cite{sevenscenes}. Our technique successfully localized all the images in the query set. Table 5 shows the median and percentile errors of our technique.

\subsubsection{Fire}
Fire dataset \cite{sevenscenes} is composed of 4000 images. It is part of the seven
scenes RGBD database. The scene consists of an area with fire equipment, chair and some other stuff. For the train and query sets, we used the standard split defined by the authors of the dataset \cite{sevenscenes}. We were able to localize all the images in the query set using our technique. Table 6 shows the different types of errors of our technique on the Fire dataset.

\subsubsection{Pumpkin}
Pumpkin dataset is one of the datasets in the seven scenes RGBD database \cite{sevenscenes}. It contains 6000 images in total of the kitchen area, along with a pumpkin in the center. We used the standard distribution of images for the training and the query sets as defined by the authors of the dataset. The percentile and the median positional and rotational errors are shown in Table 7. The proposed technique successfully localized all the images in the query set.

\subsection{Comparison with Other Approaches}
We also compared the performance of our technique with popular and state-of-the-art SfM and Network based approaches, on five of the common datasets between the different works. We only report the median location and orientation errors for comparison, as only median errors are reported by other methods. Despite the large differences between the approaches, we achieved competitive accuracies for the outdoor datasets. For the indoor datasets, we achieved the best accuracies for position estimation. Table \ref{tab:comparison} shows the median error of our technique as compared to other SfM-based and Network-based pose estimation methods.

\section{Conclusion and Future Work}\label{conclusion}
This paper has introduced a novel concept to directly compare 3D and 2D descriptors extracted using 3D and 2D feature techniques, respectively. We have demonstrated the performance of this technique for the application of 6-DOF pose localization of a query image in a dense 3D point cloud. This approach makes the localization pipeline much simpler compared to SfM or neural network based approaches. Unlike SfM-based approaches, the proposed technique can localize images in point clouds that have been directly generated from any 3D scanner. Our technique can work in both indoor and outdoor environments. Experiments show that the proposed technique achieves competitive or superior pose estimation accuracy, compared to other approaches. In the future, we will further refine our technique, followed by more detailed experimentation and demonstrate its use for other tasks that can benefit from complementing 2D with 3D data.

{\small
\bibliographystyle{ieee}
\bibliography{egbib}
}

\end{document}